# Corpus Statistics in Text Classification of Online Data

Marina Sokolova
IBDA@ Dalhousie University
& University of Ottawa
Ottawa, ON, Canada
sokolova@uottawa.ca

Victoria Bobicev
Moldova Technical University
Chisinau, Moldova
victoria.bobicev@gmail.com

**Abstract.** Transformation of Machine Learning (ML) from a boutique science to a generally accepted technology has increased importance of reproduction and transportability of ML studies. In the current work, we investigate how corpus characteristics of textual data sets correspond to text classification results. We work with two data sets gathered from sub-forums of an online health-related forum. Our empirical results are obtained for a multi-class sentiment analysis application.

## 1 Introduction

Transformation of Machine Learning (ML) from a boutique science to a generally accepted technology has increased importance of reproduction and transportability of ML studies. Diverse textual data sets complement each other and produce a fertile ground for development of new, more powerful Text Analytics methods (Fankhauser, Knappen, & Teich, 2014). On the other hand, comparison of those methods and evaluation of the results becomes more problematic as the number of the contributing data characteristics becomes intractable and in some cases contradictory. Even within the same ML task, more data does not always benefit algorithms. For example, in binary classification of ironic tweets, probabilistic and instance-based methods perform worse with more data (Charalampakis, Spathis, Kouslis, & Kermanidis, 2016). However, selection of appropriate ML algorithms is still dominated by comparison of algorithms' performance on one or several data sets (Taboada 2016).

In the current work, we focus on corpus characteristics of textual data sets and their correspondence to text classification results. We analyze how corpus statistics can be used to estimate and, consequently, improve data comparability. We work with two data sets gathered from sub-forums of an online health-related forum. Our empirical results are reported for a multi-class sentiment analysis application.

## 2 Motivation and Related Work

A common approach to invoke data in the evaluation process is to report the data source (e.g., medical records, Twitter) and its scale (e.g., the number of words/tokens, number of examples). Identifying the source implies qualitative characteristics of the data sets, such as noisiness, sampling bias, concept drift, as well as specific text characteristics, such as pragmatic use, semantic level, syntactic diversity and lexical richness (Eisenstein, 2013). The scale measures allow for direct data set comparison, especially when the size measures are supplemented by diversity measures, e.g. type/token ratio and the data label distribution. This data comparison, albeit helpful, can be further improved by estimating similarity and dissimilarity of the data sets.

Accounting for text similarity and difference has been a cornerstone of corpus linguistics studies (Kilgarriff, 2001)(Kilgarriff & Grefenstette, 2003). Degree of comparability can be evaluated by comparing whole sets, examples within the sets or on sub-example level.

Variability between data set parameters and within a single set is often estimated through word-frequency statistics. Variability estimates can be improved by increase in the level of precision with which the variability of the corpus is quantified and specifically addressing events with low frequencies (i.e., rare events) (Gries, 2006). Text can be evaluated by the degree to which grammatical features exhibit similar frequency distributions and factorial structures among the corpus parts.

In domain adaption, it is common to compare data sets through token and type estimates. Text classification results can put in perspective by computing Kullback-Leibler Divergence (relative entropy) and Cross Entropy of the textual data sets (Uribe, Urquiz, & Cuan, 2013). The measures are calculated on features representing the data sets in ML experiments. Evaluation of results in higher semantic and pragmatic Text Analytics tasks (e.g., sentiment analysis) may benefit from knowledge of domain complexity. Domain complexity reflects on difficulty of text classification for a given task. Rare words, type-token ratio and relative entropy, corpus self-similarity (e.g., Jensen-Shannon divergence) can be used to evaluate domain complexity (Remus & Ziegelmayer, 2014).

Corpora similarity and domain complexity are essential in the case of manual annotation of sentiments. Multi-class sentiment annotation of two corpora has been compared in (Bobicev & Sokolova, 2017). Both corpora were built from texts gathered from the same online forum. Each text was annotated by three annotators. The inter-annotator agreement was Fleiss kappa=0.46.

## 3  Lexical Richness Measures

Independent variables of the corpus statistics are based on count of all words (tokens), different words (types) and their occurrence in the corpora: $N$ is the text size, or the number of all the words found in the text, $V$ is the vocabulary size, or the number of different words in the text, $m$ is the number of times a given type occurs in the corpus (Oakes, 2005).

In many cases, unique data characteristics can be found by analysis of out-of-vocabulary (OOV) words. OOV words mostly appear among *hapax legomena* (i.e., types occurred once in the data, $m = 1$) and *dis legomena* (i.e., types occurred twice in the data, $m = 2$). Those rarely occurred words are more frequent in text posted on online public forums and social networks; those texts are less contrived and poorer edited than mainstream media texts.

Another set of statistics measures estimates lexical richness of corpus. A type/token ratio $\frac{V}{N}$ is one of the basic corpus measures. The ratio approximates vocabulary richness. A higher richness makes text classification more difficult for automated analysis; hence the same $F - score$ will signify more insightful and accurate classification if obtained on text with a richer vocabulary than with a scanty one.

$\frac{V}{N}$ can be enhanced through vocabulary frequencies' distribution $\frac{V(m,N)}{V}$, where $m$ is the number of times the type appears in the corpus. For a given corpus, empirical $\frac{V(m,N)}{V}$ indicates how well the text is contrived.

A more contrived text has more mid-range frequencies and fewer low-range frequencies. $\frac{V(1,N)}{V}$ and $\frac{V(2,N)}{V}$ evaluate ratio of rare words in vocabulary; $\frac{V(1,N)}{V}$ is often used as a proxy of noiseness of a corpus.

Note that even texts collected from the same source may exhibit different characteristics. For example, online forums can be affected by shifts in authorship demographics (Bobicev & Sokolova, 2015), emergence of new topics due to external events (advances in science, political events, or natural disasters). All those changes increase variety in posted text. At the same time, knowing diversity among corpora can yield better retrieval/identification of a positive class (Aly et al, 2014).

To look for unifying characteristics among corpora, we consider lexical and content-focused statistics. Lexical density is the ratio of content words (nouns, verbs, adjectives) to the number of all words. It is the percentage of lexical as opposed to grammatical items in a given text or corpus of texts. Lexical items are subjects to a much slower change than grammatical items, thus better represent continuity and coherence in a corpus. Lexical density can be estimated by $\frac{V(m,N)}{V}$, where $m$ is mid-range for the analyzed corpus. Mid-range $m$ associate with content words (i.e., terms that define and contribute to the text content); higher mid-range $m$ signify a bigger information load of the text.

## 4  Empirical Study

### 4.1  Data sets

We experimented with two data sets collected from sub-forums of a health-related online forum. The data set A consisted of texts posted by 359 authors, the data set B – by 355 authors. Among those authors, 133 authors had messages in both sets, albeit their contributions were asymmetrical: they had several posts in one set and fewer messages in the other set.

Previously, the data sets were manually labeled for a multi-class sentiment analysis study (Bobicev and Sokolova, 2017). In the current study, we removed posts on which the annotators disagreed and worked with the remaining unambiguous posts. Table 1 reports parameters of the two data sets.

Table 1. Data set parameters

| Parameters | Set A | Set B |
|---|---|---|
| Authors | 359 | 355 |
| authors contributed to one data set | 226 | 222 |
| authors contributed to both data sets | 133 | 133 |
| Topics | 80 | 64 |
| all posts | 1321 | 1000 |
| post per topic | 16.5 | 15.6 |
| post per author | 3.7 | 2.8 |
| ambiguous posts | 337 | 30 |
| unambiguous posts | 984 | 970 |
| unambiguous post per topic | 12.3 | 15.2 |
| unambiguous post per author | 2.7 | 2.7 |

**4.2 Corpus Statistics**

We built two corpora from the data sets (Corpus A and Corpus B respectively). As corpus statistics is based on word count, tokenization is an essential operation of the corpus construction, e.g., different tokenization procedures produce different statistics on the same data set. For example, word capitalization can be left "as is" or transformed to lowercase. In online data, texts often include non-lexical tokens (dates, URL, money amount). We decided to replace all non-alphabetical characters by spaces and separate remaining words. We also transformed all letters in lowercase.

To approximate an upper bound of middle-range *m*, we found the most frequent topic-related noun (*clinic* in the two frequency lists). All the types that had a higher frequency corresponded to stop and short words. To approximate a lower bound, we looked at types with m=1,2,3,4. Types that occur 4 times contained topic-related nouns, adjectives, and verbs; hence, we counted *m*=4 among middle-range m. The resulting *mid m* reports the number of word types that occur

> 3 times after removal of the stop words types. The descriptive statistics of the two corpora is presented in Table 2.

For both corpora, the length of posts in words closely follows Zipf's law, as shown on Figure 1: there are a few quite long posts and many short messages.

Table 2. Statistics of the two corpora.

| Parameters | Corpus A | Corpus B |
|---|---|---|
| words (tokens) | 120 077 | 108 245 |
| different words (types) | 6 375 | 6 297 |
| number of sentences | 8 853 | 7 973 |
| words per post | 122 | 112 |
| sentences per post | 9.0 | 8.2 |
| words per sentence | 13.6 | 13.6 |
| *mid m* | 2105 | 1999 |
| m = 1 | 2738 | 2870 |
| m = 2 | 988 | 904 |
| m= 3 | 491 | 448 |

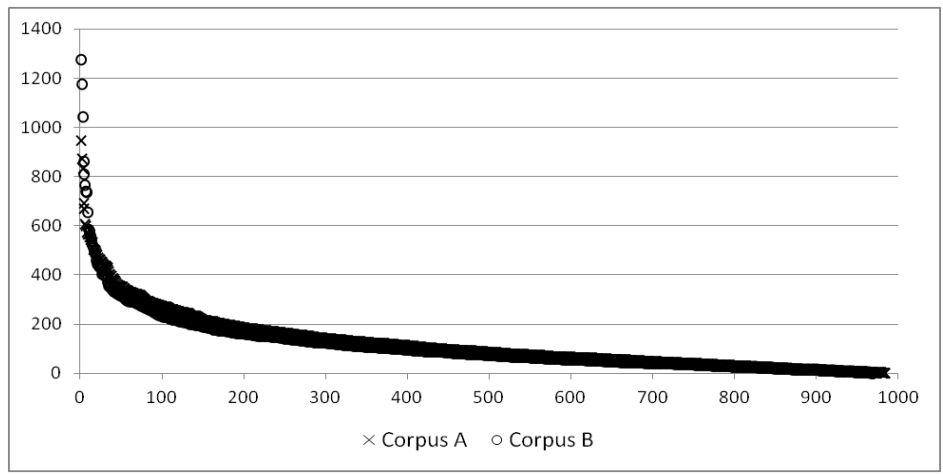

Figure 1. The length of the posts measured in words. The posts are sorted by their length. Axis X shows posts, axis Y shows numbers of words in the post.

### 4.2 Vocabulary richness

To evaluate vocabulary richness of the corpora, we computed measures discussed in Sec 3. Table 3 reports the results. We used two-tailed, non-paired t-test to assess statistical difference of the results; the two-tailed, non-paired version was used as the values were obtained on two different corpora. The test indicated that difference between the five measures is not significant (P value = 0.9981); hence, the basic characteristics of the corpora are compatible. At the same time, $\frac{V(mid\ m,N)}{V}$ is higher for Corpus A than for Corpus B. This lexical density comparison indicates that Corpus A has a richer lexical vocabulary and consists of more informative texts than Corpus B. On the opposite end of the frequency spectrum, $\frac{V(1,N)}{V}$ is higher for Corpus B than for Corpus A, thus Corpus B has a higher noise than Corpus A. Those results complement each other, as information-dense content tends to have less noise.

Characteristics of information-dense texts (in our case, Corpus A) include semantically related nouns (domain terms), a limited set of qualifiers. Those texts use a few selected grammatical constructions (Fankhauser, Kermes, & Teich, 2014). In contrast, lesser information density (in our case, Corpus B) corresponds to topic elaboration, a frequent use of co-references, conjunctions, auxiliary and modal verbs; hence such texts are prone to grammatical and spelling mistakes.

Table 3: Lexical statistics of the two corpora

| Measures | Corpus A | Corpus B |
|---|---|---|
| $\frac{V}{N}$ | 0.05309 | 0.05817 |
| $\frac{V(mid\ m,N)}{V}$ | 0.33020 | 0.31745 |
| $\frac{V(1,N)}{V}$ | 0.42949 | 0.45577 |
| $\frac{V(2,N)}{V}$ | 0.15498 | 0.14356 |
| $\frac{V(3,N)}{V}$ | 0.07702 | 0.07115 |

# 5 Text Classification Experiments

In our study we represented texts through a) Bag of words (BOW) features, b) sentiment-bearing features, and c) features selected from the two feature sets:

a) to build BOW representation, we combined all the posts from both corpora, created the frequency dictionary and removed all words with occurrence 1. This, we obtained 5784 features;

b) to extract sentiment-bearing features, we used popular sentiment resources: SentiWordNet, Bing Liu Sentiment Lexicon, SentiStrength, AFINN Hashtag Affirmative and Negated Context Sentiment Lexicon, Sentiment140 Lexicon, MPQA, DepecheMood, Word-Emotion Association Lexicon, General Inquirer, and HealthAffect. We commenced all the lexicons, removed duplications and removed terms that did not appear in the posts. We had 10541 features. The number of features is much larger than in the BOW representation due to collocations and word combinations contained in some lexicons;

c) we created the third feature set by combining BOW and sentiment features and applying the Best Subset feature selection: 457 features.

## 5.1 Sentiment classification

To assess how the corpora characteristics influence text classification results, we use a sentiment classification task. We classify text into four classes: *confusion*, *encouragement*, *gratitude*, *facts*. The class distribution in Corpus A: *confusion* - 117 posts, *encouragement* - 310 posts, *gratitude* - 124 posts, *facts* – 433. The class distribution in Corpus B: *confusion* - 146 posts, *encouragement* - 494 posts, *gratitude* - 69 posts, *facts* - 261.

We differentiated between quantifying effects of corpora in classifiers' training phase and in classifiers' test phase. We constructed 2x2 sets of Machine Learning problems, with each corpus being a training set in two problems and a test set – in two problems. Namely, in one problem we trained classifiers on the complete Corpus A and then tested the classifiers on the complete Corpus B. In another problem, we used 10-fold cross-validation on Corpus A to train and test classifiers. Similarly, we trained classifiers on Corpus B and tested them on Corpus A in another problem. In the last setting, we applied 10-fold cross-validation to train and test classifiers.

We used the majority class baseline for each problem. Table 4 reports on the baseline *F-score (F)* and corresponding *Precision (Pr)* and *Recall(R)*.

Table 4: The majority class classification baselines.

| Evaluation | F | Pr | R |
|---|---|---|---|
| train on Corpus A, test on Corpus B | 0.114 | 0.072 | 0.269 |
| Corpus A, cross-validation | 0.269 | 0.194 | 0.440 |
| train on Corpus B, test on Corpus A | 0.151 | 0.099 | 0.315 |
| Corpus B, cross-validation | 0.344 | 0.259 | 0.509 |

We applied multi-learning versions of Naïve Bayes, DMNBtext, NBMultinomial and SVM algorithms embedded in the Weka toolkit[1]. Table 5 reports the highest F-score (*F*) and corresponding Precision (*Pr*) and Recall *(R)* when training is done on Corpus A; Table 6 reports the results when on Corpus B.

Table 5. Sentiment classification results: training on Corpus A.

| Features | Evaluation | F | Pr | R | Algorithm |
|---|---|---|---|---|---|
| BOW | train on Corpus A, test on Corpus B | 0.473 | 0.639 | 0.484 | DMNBtext |
| Lexicons | train on Corpus A, test on Corpus B | 0.510 | 0.549 | 0.509 | SVM |
| Selected | train on Corpus A, test on Corpus B | 0.537 | 0.621 | 0.539 | DMNBtext |
| BOW | Corpus A, cross-validation | 0.628 | 0.629 | 0.639 | DMNBtext |
| lexicons | Corpus A, cross-validation | 0.641 | 0.644 | 0.652 | DMNBtext |
| selected | Corpus A, cross-validation | 0.744 | 0.745 | 0.745 | NBMultinomial |

---

[1] https://www.cs.waikato.ac.nz/ml/weka/

Table 6: Sentiment classification results: training on Corpus B.

| Features | Evaluation | F | Pr | R | Algorithm |
|---|---|---|---|---|---|
| BOW | train on Corpus B, test on Corpus A | 0.475 | 0.607 | 0.506 | DMNBtext |
| Lexicons | train on Corpus B, test on Corpus A | 0.525 | 0.604 | 0.547 | DMNBtext |
| Selected | train on Corpus B, test on Corpus A | 0.565 | 0.632 | 0.579 | NBMultinomial |
| BOW | Corpus B, cross-validation | 0.560 | 0.568 | 0.592 | DMNBtext |
| lexicons | Corpus B, cross-validation | 0.573 | 0.576 | 0.598 | DMNBtext |
| selected | Corpus B, cross-validation | 0.630 | 0.632 | 0.634 | NBMultinomial |

## 5.2 Discussion of the results

The classification results show that training classifiers on a less-noisy, more informative corpus brings a partial advantage over training the same classifiers on a less informative and content-oriented corpus. The positive impact of a more informative corpus is apparent when we compare the results of two 10-fold classification evaluations. In those problems the classifiers are tested on the same corpus as they were trained on; see the lower part of Table 5 for the results on Corpus A and the lower part of Table 6 for the results on Corpus B. If we evaluate difference in *F-score*, the two-tailed P = 0.1216 indicates a near-significant difference. After we extend statistical generalization to *F-score* and *Precision*, the classification results of Corpus A and Corpus B become significantly different (the two-tailed P = 0.0117). We hypothesize that an information-dense and content-dense Corpus A provides advantage in correct identification of positive examples through a higher *Precision*. This can be explained through a learning bias, when learning algorithms often rate previously unseen test examples based on similarity of their text representations to those of the positive training examples, striving to achieve a higher *Precision*.

When the classifiers are trained on Corpus A and then tested on Corpus B, they obtain lower *F-score* than in the reverse learning scheme (i.e., the classifiers are trained on corpus B and then tested on corpus A), although *F-score* difference is not statistically significant, with the two-tailed, non-paired P = 0.6633. The achieved *Recall* values show that training classifiers on Corpus B while testing them on Corpus A considerably improves extraction of positive examples if compared the reverse learning scheme (i.e., training the classifiers on Corpus A and testing them on Corpus B), although this improvement still falls short of significant improvement, with the two-tailed, non-paired P = 0.2759. We hypothesise that a greater grammatical and vocabulary diversity in Corpus B allows algorithms to build a wider representation for positive examples during the training phase, thus capturing more positive examples during the test phase.

## 6 Conclusions and Future Work

In the current work, we have studied correspondence between corpora statistics and text classification results. We have used multi-class sentiment classification as the empirical setting for four evaluation schemes. We have shown that statistical characteristics of the corpora and assessment of corpora similarity and differences are essential in evaluation and understanding of text classification results. As data sets available for Text Analytics become more diverse, corpus analysis becomes a necessary step in estimation of algorithms' performance on new data sets.

For future work, we consider to integrate corpora characteristics in classification evaluation. Direct reporting on similarity and dissimilarity of the data can considerably improve efficiency of Machine Learning algorithms. For example, we can separate the results obtained on examples similar for the data sets from the results obtained on examples on which the data sets differ. This requirement corresponds to condition of *similarity of algorithms*, which expects that similar instances should be classified similarly (Andersson, Davidsson, & Lindén, 1999).